# Using Clustering to extract Personality Information from socio economic data.

Alexandros Ladas, Uwe Aickelin, Jon Garibaldi
Department of Computer Science
University of Nottingham
Nottingham, UK
ayl@cs.nott.ac.uk, uwe.aickelin@nottinham.ac.uk,
jmg@cs.nott.ac.uk

Eamonn Ferguson
School of Psychology
University of Nottingham
Nottingham, UK
eamonn.ferguson@nottingham.ac.uk

*Abstract*— **It has become apparent that models that have been applied widely in economics, including Machine Learning techniques and Data Mining methods, should take into consideration principles that derive from the theories of Personality Psychology in order to discover more comprehensive knowledge regarding complicated economic behaviours. In this work, we present a method to extract Behavioural Groups by using simple clustering techniques that can potentially reveal aspects of the Personalities for their members. We believe that this is very important because the psychological information regarding the Personalities of individuals is limited in real world applications and because it can become a useful tool in improving the traditional models of Knowledge Economy.**

## I. INTRODUCTION

It is now a common belief that Knowledge Economy has a lot to gain by including psychological information regarding the Personality of individuals into its methods and practices. Personality Psychology's rich theory provided in an appropriate processed form can enrich models that are applied in Economics, including models that come from the fields of Machine Learning and Data Mining. To be able to utilize psychological information regarding the Personalities of the individuals in order to enhance Data Mining methods, the details about the Personality must be either provided in the appropriate form or extracted directly from the data and processed properly. Our purpose in this work is to suggest a new way to extract psychological characteristics of Personality from alternative sources of data by using clustering algorithms. We consider this a very important task not only in cases where this kind of psychological information is not provided, which is very common in the data gathered from real world applications, but because it also provides a solid proof that economic data have the potential to reveal characteristics about the Personality and thus using them in order to improve the accuracy of Data Mining models in economy is unbiased and valid.

The importance of accounting individual differences in economics is highlighted in very diverse areas in bibliography. From Personality Psychology and Economics' perspective [4] it is emphasized that information regarding the Personality, when applied in a theoretically meaningful way, can provide further insight in interpreting complexities and patterns of economic behaviour. From Computer Science perspective, particularly in Behavior Mining as it is defined in [11], the procedures of Data Mining and Machine Learning must take into consideration the semantics of the data, which within the economic context can be Personality characteristics that govern the transactional data.

Therefore our aim is to build a model which utilizes psychological characteristics of Personality in order to guide Data Mining techniques in extracting valuable knowledge that can explain economic behaviours. Central role in building this model plays the notion of behaviour. In [6] behaviour is defined as the action or reaction of an entity to stimuli in environment. Since in Personality Psychology, behaviours are described as manifestations of Personality traits, the task of mining behaviours from economic datasets that will potentially lead in identifying psychological characteristics of Personality seems to be essential and integral part of our procedure. These characteristics can help us sketch the Personality Profiles of individuals which can be further used to enhance existing Data Mining Models.

The general plot of our model is depicted in Fig. 1. It can be seen that at first behaviours are being extracted from the economic data and then they are used to sketch the Personality Profiles of the individuals. Then the Economic data together with the Profiles are used to enrich the existing economic data mining process in order to discover complicated patterns of economic behaviour. The role of economic data here is twofold, posing as the traditional framework in which economic Data Mining is performed and as an alternative source of extraction of Personality's characteristics, in case the latter are not available.

In this paper we propose a method of implementing the part of our model that is related with the mining of Behavioural Groups by utilizing traditional Data Mining techniques. More specifically we perform clustering on a socio economic dataset provided by the Consumer Credit Counseling Service which includes demographic information and financial, debt and expenditure details. Our intuition is that information about how clients decide to spend their money together with their financial and debt details hold significant information about the

---

This work is partially funded by Experian

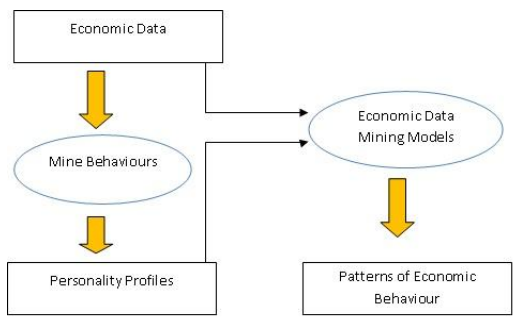

**Figure 1 model of incorporating personality information into data mining methods**

preferences of the clients and that with the appropriate procedure we may discover different behaviours among users. Our goal is to discover these different behaviours by using traditional clustering algorithms to cluster clients in different Behavioural Groups. In addition to this, we provide a novel method of Personality Profiling from alternative data sources by utilizing the ratings of Selfishness, provided by 52 Psychology students for expenditure attributes of the dataset to characterize the Behavioural Groups.

We show that mining behaviours is possible by using simple data mining techniques to partition people regarding their expenditure and eventually identify patterns in people's expenditure. Furthermore we propose a simple method to associate these Behavioural Groups with the Selfish or non Selfish aspects of Personality. Despite its drawbacks our work reveals a novel way to extract psychological characteristics of Personality through static socio economic data. The findings of this work can encourage more sophisticated research on Personality Profiling through mining socio economic data and in a greater extent using these profiles in order to power up our economic Data Mining models in extracting and interpreting patterns of economic behaviours more accurately and precisely.

The rest of the paper is organized as following. In section 2 the related work is discussed, in section 3 the details of our data and selfish scores are presented and in section 4 we describe our proposed framework for Behavioural Group mining in detail, proposing a way to characterize the clusters as markers of Selfish or non Selfish behaviour. In section 5 we analyze the results of our Cluster analysis and Selfishness characterization. Finally in section 6 we conclude our work and we present our thoughts about the future steps of our research.

## II. RELATED WORK

Similar work to our effort in extracting personality information from economic datasets exists in bibliography in many Psychology and Economic articles. The majority of them though adopt a procedure of extracting Personality Information that includes additionally information that is gathered through questionnaires and experiments with well known economic games. Our approach on the other hand tries to extract behaviours that can be associated with Personality traits through the application of data mining techniques exclusively on economic data without using this information that is rarely given in datasets from real world applications. A very interesting and innovative overview of this work is provided by

[1] and [4] where they present models to measure Personality within economic context and they show how the models from Economics can benefit from the models of Psychology and vice versa. In particular, in [4] the authors state that the economic modeling of preferences would benefit from the application of psychometric principles and that Personality Information can provide alternative ways of research in Economics, introducing a theoretical background of explaining anti-social traits, an area of Personality Psychology that was undermined by traditional economic analyses. Although the models that are discussed are very different from the perspective of data mining, their results can be used to provide a better insight of the data and guide our data mining procedures or even enhance them.

From a computer science perspective, Behaviour Informatics defined in [6], offer a way to mine behaviours from the data in order to provide a deep understanding of behaviours. In this work a new framework is proposed in order to transform the data properly and then a general algorithm is provided in order to extract behaviours. In addition to this the significance of this in business intelligence and in credit estimation is highlighted. The proposed framework is very complicated and it can be applied mainly on activity data, where an action can lead to a result and thus a relationship with goals and motives can be defined. This is very different from the case of socio economic data as the information that is available there does not include actions and objects of the actions. However the idea behind this approach can be very useful, especially in the future of my research, as it provides a method of transforming the existing data into similarly behavioural data by including into them, principles from the theory of Personality Psychology.

The same basic idea is demonstrated in [10] where a mining approach similar to A-priori is performed in stock market data in order to find peculiarity groups. This is achieved by first transforming the stock market data in day by day behavioural datasets and then applying a mining algorithm on this data. The novelty of this work can give an example of how to transform the data in appropriate way in order to power up the procedure of mining the data. However, the stock market data are similar to activity data and thus this idea cannot be applied in the same way to our data. Besides, their application is very specific whereas our model aims to be more general and can be applied in more datasets and not only in stock market data. A very interesting work is described in [9], where an innovative procedure is introduced, that performs data mining upon credit rating data in order to classify credit holders in good bad. Their work utilizes the nearest subspace method in order to achieve a 2 class classification.

Finally, the field of Behavior mining introduced in [11] can provide some interesting and innovative ideas about how we can proceed with the construction of our model, depicted in Fig 1. In more detail, [11] defines a new field of Data Mining, namely Behavior Mining, in which the semantics of the data should be taken into consideration. By taking advantage of the semantics encapsulated within the data we will be one step closer to achieve our goal in discovering valuable knowledge in economics. Therefore, by including principles from Personality Psychology as semantics to our Data Mining techniques can help us improve our analysis in the field of Economy. Although as a goal of our research is to build a model that will combine additional information that derives from the principles of Personality Psychology with Data Mining techniques and methods, our difference with the Behavior Mining approach is that we are unsure whether this information should be necessarily included as semantics to our data. Moreover we think that the proposed framework of behavioural data in [11] to encapsulate the semantics in our data, increases the complexity of the analysis and it cancels out traditional and powerful methods of Data Mining which perform on simple data and not complex.

Despite the fact that all these Behaviour mining methods present innovative ideas, especially for the future part of our research they suffer from certain limitations. First of all, they are data specific, being able to be applied exclusively in activity data and not on other kinds of data. In addition to this, the mining of Behaviours in these models is not being presented as part of Personality Profiling process and neither they are associated with Personality Psychology. Therefore, our method of mining Behaviours from socio economic data and associating them with aspects of Personality Psychology poses as a novel process of Behaviour mining and Personality Profiling on alternative data sources.

## III. DATA PRESENTATION

### A. Data Analysis

The CCCS dataset, introduced in [3], is a socioeconomic crossectional dataset based on the data provided by the Consumer Credit Counseling Service. In its 58 attributes it contains information about 70000 clients who contacted the service between the years 2004 and 2008 in order to require advice about how they can overcome their debts. The information was gathered through interviews when each client first contacted the service and it varies from standard demographics to financial details, expenditure details and debt details.

Based on this fact we can see that the 58 attributes of the dataset are organized to more general categories of attributes like the above. They form a sort of hierarchy which can be seen in Fig. 2. Here the CCCS dataset is splitter into 6 big categories, Demographics, Debt, Assets, Expenditure, Debt Details and Income. This is similar with the above with the only difference being that Financial details are split into Income and Assets and the information about Debt is split into Debt Details and Unsecured Debts. Because of the importance Debt and Income possess, it was better to be viewed separately in this abstraction and not as part of other categories. That is the reason that attribute Income is linked directly to the CCCS dataset. The rest of the attributes falls into the 5 subcategories of the CCCS dataset. The description of the important attributes of the dataset is given in the table1.

### B. Data preprocessing

Before moving to the cluster analysis of the data, it is always important to analyze it, find inconsistencies within the data and offer a way to deal with them. In CCCS dataset there were a lot of issues to address before clustering the data regarding missing values, nominal attributes, duplicates, correlations and a decision to make about whether or not we should take into account the time parameter. The issues that came to our notice and the actions we took in order to deal with them are the following:

- **Missing Values.** Although the percentage of missing values is around 18% of the data, they tend to appear in the same clients for the same type of categories. With this in mind we were able to delete the specific clients with regard to the attributes we were interested and move to a smaller dataset with a very small percentage of missing values. In the case of using all attributes we were able to reduce the size of the dataset to approximately 37000 with a percentage of 0.8% missing values. This small proportion of missing values gave us the opportunity to simply replace them with the mean in case of numeric attributes or the mode for the nominal without worrying about the impact on the cluster analysis as their number is very small.

- **Duplicates.** In the CCCS dataset it was noticed that there were entries with the same client id. These clients, approximately 5000, were removed because they would be considered as noise to our further analysis.

- **Time parameter.** The CCCS dataset contains the attributes regarding the month and the year the client contacted the service to seek advice. We decided to disregard the time parameter from our analysis since the date of contact is irrelevant to the purposes of our analysis, assuming that the characteristics of clients are stable over time. Although this assumption is valid for almost all attributes except debt, which as stated in [3] is being reduced in a steady rate over time, we decided to not to consider the time parameter after all because it needs a time series analysis which is very different from the purposes of this work.

- **Correlations.** Using the Pearson's coefficient all the correlated pair of attributes were identified and each attribute of each pair with coefficient more than 0.95 or less than -0.95 was removed as it does not contribute anything more than its paired attributes. So attributes like mortdebt and the cpcat - cpsc were removed as they were correlated to hvalue and udcat – udsc respectively.

- **Nominal Attributes.** Despite the fact that most of the attributes within the dataset are numeric there are 6 attributes that fall in the demographic category that are nominal. Due to the fact that clustering requires the data to be in numeric format in order to be applied we decided to transform these attributes to numeric by simply assigning a numeric value to each type. However treating nominal values as numeric is just an oversimplification while these two types of data hold significant differences. A better way would be to transform in an appropriate way as numeric in order to contribute valuable information to our further analysis.

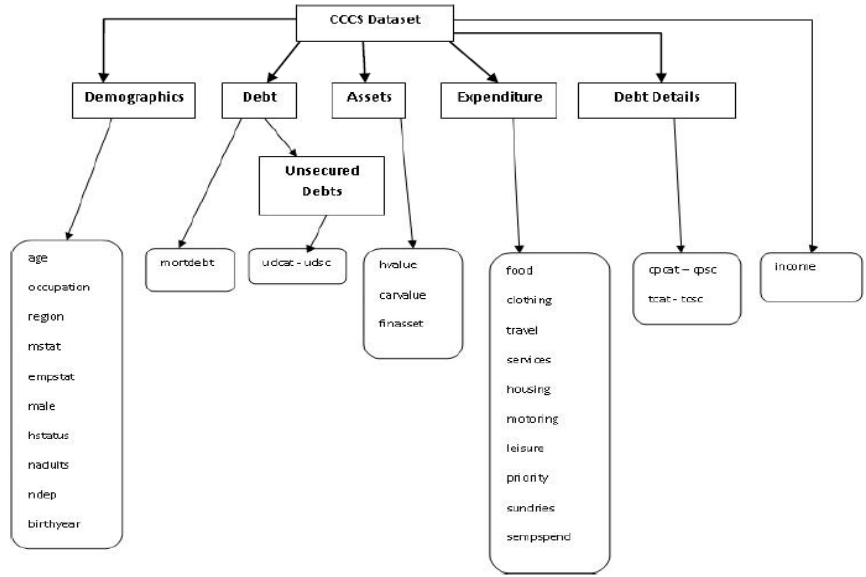

**Figure 2 The structure of CCCS data**

- **Occupation Attribute.** Lastly although this attribute is considered to contain valuable information which is often associated with Psychological Information we had to remove it from our cluster analysis as it contains a very large number of missing values, nearly half of the size of clients. Similarly with the previous case a good way to transform this in a numeric attribute that will contain valuable information is required, because this attribute is also found among the nominal ones.

### C.  Data Description of CCCS attributes

In the Fig. 3 and Fig. 4 we can see the boxplots of type of financial, debt and expenditure attributes. These boxplots provide a better insight of the data of CCCS revealing important information about the statistics of these attributes. The first thing we can notice in these boxplots is the extremely large magnitude of attributes like hvalue, udebt and mordebt and the relatively large magnitude, compared to their type's values, of income and housing. These attributes will play an important role in our Cluster analysis because of the large magnitude of their values. Something that we should notice is that they may contain outliers like houses with value of 300000 pounds as we can see in the Fig. 3. These observations are very crucial for the cluster analysis that it will be conducted on this dataset as it will help us explain the output of the clustering methods.

### IV.  CLUSTERING FRAMEWORK

Since our aim is to discover Behavioural Groups of clients in this dataset taking advantage of the information that is enclosed in expenditure types, financial and debt details, we need a clustering method that will produce clusters that are well separated regarding these attributes and that will potentially define groups of people with different habits,

**Table 1 Description of the important attributes**

| Variables | Description |
|---|---|
| pid | individual identifier |
| udebt | total value of unsecured debt |
| mortdebt | total value of mortgage debt |
| hvalue | total value all housing owned |
| finasset | total value of financial assets |
| carvalue | resale value of car |
| mortterm | months left to pay on mortgage |
| clothing | total monthly spending on clothing |
| travel | total monthly spending on travel |
| food | total monthly spending on food |
| services | total monthly spending on utilities |
| housing | total monthly spending on housing |
| motoring | total monthly spending on motoring |
| leisure | total monthly spending on leisure |
| priority | total monthly spending on priority debt |
| sundries | total monthly spending on sundries |
| sempspend | total monthly self-employed spending |
| other | total other spending |
| income | total monthly income |
| udcat - udsc | variables recording balances on unsecured debts for: catalogues, collection agency, credit card, ge capital, overdraft, personal loan, store card, other |
| cpcat-cpsc | variables recording contracted monthly payments for each of above, where applicable |
| tcat-tcsc | term remaining on repayments for each of above, where applicable |

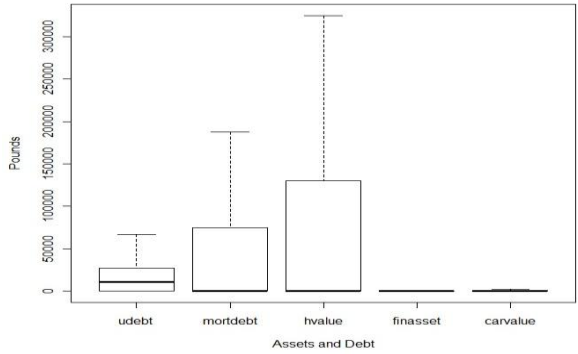

**Figure 3 Boxplots of Assets and Debt**

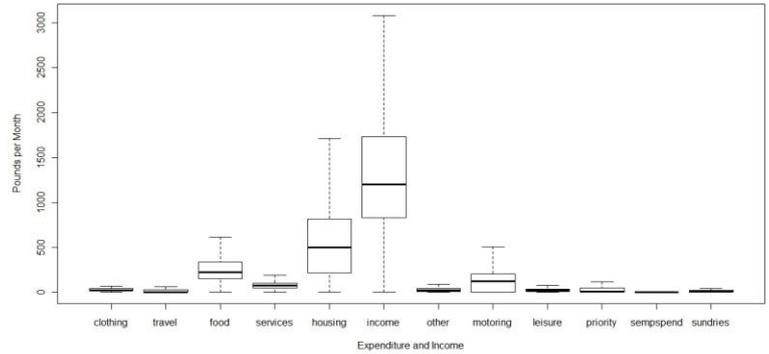

**Figure 4 Boxplots of expenditure and income**

preferences and characteristics.

For this reason we decided to apply traditional partitioning clustering algorithms on this dataset. Due to the huge size of the CCCS dataset our choices were limited to the application of K-means and Clara, which were the only partitioning clustering algorithm that could be executed in order in the R environment where the clustering analysis took place.

K-means partitions the observations into clusters based on the distance of each observation from the center of the cluster. Further information of this algorithm can be found in [7]. Clara on the other hand, introduced on [5], is appropriate for clustering large applications. It is a modification of PAM algorithm that performs clustering on a sample of data. In similar way with PAM it uses the dissimilarity matrix in order to cluster similar observations around medoids, which here are cluster members and not the centre of the clusters.

Both Clara and Kmeans require the number of clusters and the initial medoids or centroids respectively as input. That means that in our case, where the optimal number of Behavioural Groups is not known beforehand, we should perform an extensive search in order to find the optimal clusters that are well separated and compact and at the same time partition the clients with regards their expenditure types. Our intuition was that finding the optimal clusters that partition the clients in the best way, will help us detect well defined Behavioural Groups.

For this reason, we designed a process of approaching optimality, which is visualized in Fig 5. As it can be seen we experimented with different subsets of attributes stages, namely stages A, B and C that are depicted in the 3 boxes in the Fig. 5, and for each stage of experiments, we performed Kmeans and Clara for given clusters starting from 2 until 20. Then we utilized the Silhouette criterion [8] for the case of Clara and Calinski Index [2] for the case of Kmeans to help us identify good clusterings, since both of them return a value indicating how well separated and compact the clusters are. However since the best values of Silhouette and Calinski index do not necessarily indicate the optimal numbers of clusters we used the visual inspection of clusters by boxplots and biplots in order to analyze the outcome of the clustering and gain a better understanding of the clusters are defined. The extraction of Behavioural Groups was based on the agreement of all the evaluation measures we used. Finally by inspecting the values of Silhouette and Calinski, combined with our visual analysis we were able to get an indication about what the optimal number of clusters would be.

The framework was implemented in R with the help of built in functions. In order to overcome the random results obtained by the initial distribution of centroids and medoids, as Kmeans' and Clara's outcomes are dependant in great extent to the initialization of their centroids and medoids, we iterated both algorithms 100 times for each given number of clusters and we kept the best results each tune. However we have to be careful with the results of Kmeans because the algorithm is susceptible to outliers and extreme values and our dataset contains few of them that were revealed during the Data Description in the previous section.

Although our framework is a simplified approach in means of finding the optimal clusters, it provides a good estimation of the optimal clusters by utilizing, validation indices and visual inspection. This way it achieves discovering well defined clusters, which is the medium to discover well separated Behavioural Groups.

Finally, having extracted the Behavioural Groups we used the Selfishness ratings provided by 52 students of the Psychology department of University Of Nottingham who rated each Expenditure attribute with two scores from 1 to 7 in order to associate the extracted Behavioural Groups with the Selfish aspect of Personality. The first score indicates how strongly associated is the attribute with Selfish personality and the second one indicates how strongly is the attribute associated with Non Selfish Personality.

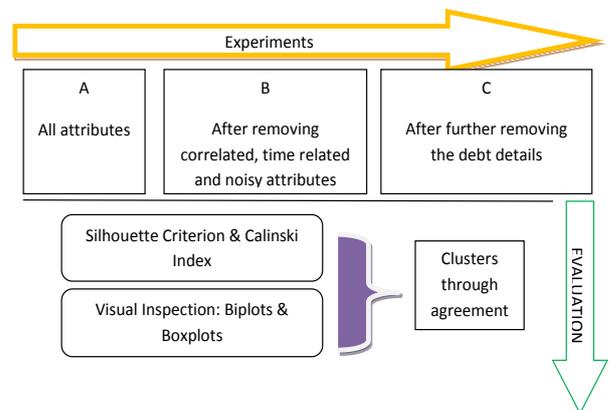

**Figure 5 Methodology of finding the optimal clusters**

## V. RESULTS

### A. Biplots

The results of applying the clustering framework on the three stages are depicted in the biplots of Table 2. The biplots are the depiction of clusters on the space that is created by the first two components of Principal Component Analysis. The clusters that are depicted are the optimal that were selected by the Silhoutte Criterion for the case of Clara and the Calinski Index for the case of Kmeans.

In case of all attributes (Stage A) Clara returned 1 big cluster and 5 smaller and non distinct. In similar way we can see that Kmeans produced 1 big distinct cluster (blue) and 6 smaller and non distinct. Within the 6 smaller clusters are included 2 clusters with just one member proving Kmeans weakness against extreme values and outliers. Similar clusters were observed in all outcomes of Kmeans in experiments of stages B and C.

When we remove the unwanted attributes defined in Stage B, Clara continues to produce one big cluster and 3 smaller that are quite more distinct than before whereas Kmeans returns 9 clusters 3 of them (blue, green and ping) are well separated.

Finally in stage C, after the further removal of debt details Clara returns 2 well distinct clusters whereas Kmeans produces

13 clusters, of which one is big and well separated.

Removing further attributes and categories of attributes than those defined in stages A,B,C in Fig. 5, resulted in Clara being stabilized at returning the same two clusters as the optimal like it did in stage C and Kmeans producing a large number of clusters as optimal. That means that the attributes removed after the stage C caused the Kmeans and Clara to stick to extreme values when proposing the optimal clusters.

Analyzing these results we can understand that the best clusters can be found within the B and C experiments. Furthermore Kmeans' weakness against extreme values and outliers which was validated in all our examples can be handled with the scaling of the data. Clara on the other hand does not need such measure. Moreover, we can spot that in all our results there is a clear big and distinct cluster. Unfortunately we are unable to check the expression of the attributes in the big cluster as well in the rest of the clusters depicted by biplots as the majority of the clusters are so small and no safe example can be made from the biplots. That is the reason why we have to check the boxplots of these attributes and especially of the expenditure and financial attributes to see if they are overrepresented in specific dusters. Finally by inspecting the values for each case of clusters returned from the application of clustering framework on all the results from the experiments we could infer that the best number of clusters lies between the cases of 4 and 6.

### B. Inspection of Boxplots

By studying the boxplots of each attribute for the clusters returned from the above clusterings we were able to spot

**Table 2Biplots of the clusters returned by Clara and Kmeans for stages A, B and C**

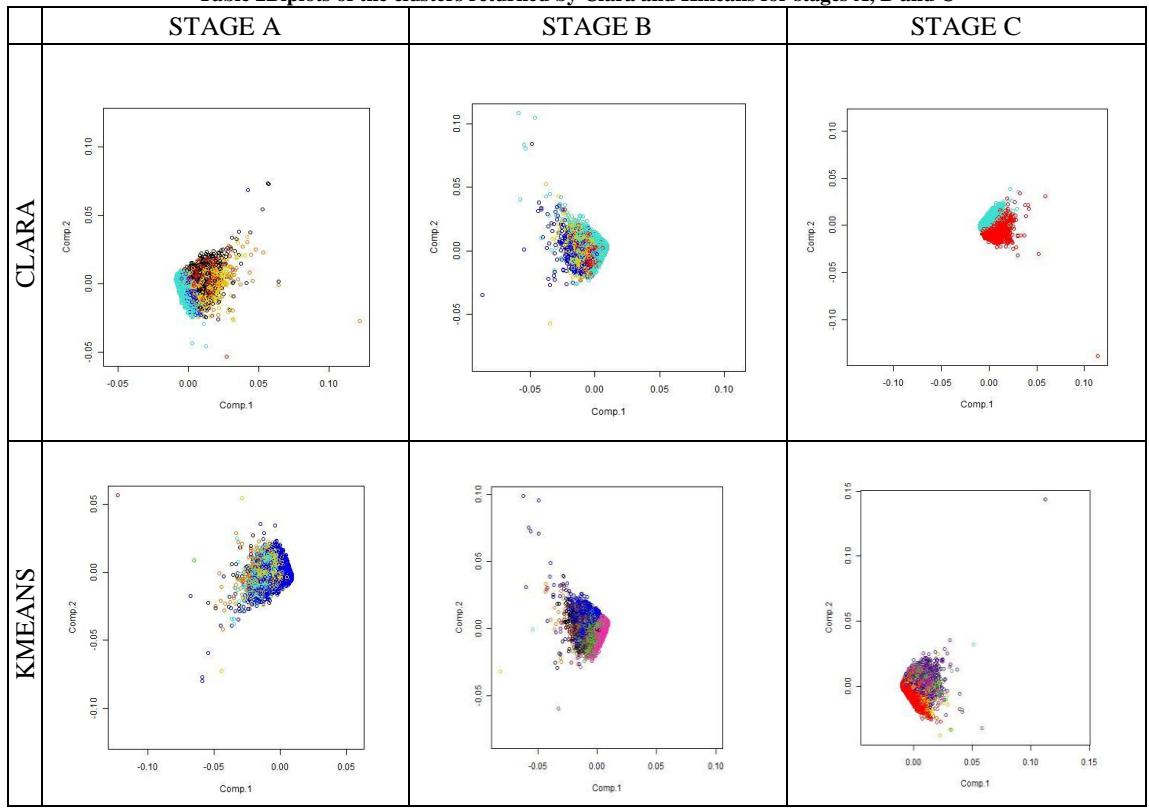

which ones are overrepresented inside the clusters and which are underrepresented. The results are aggregated in the tables 3, 4 and 5 where for each cluster the attributes that are overrepresented and the attributes that underrepresented are being indicated by the (+) the (-) sign respectively. Studying carefully these results by checking the attributes that characterize each cluster, we can see that that there are some common clusters among the different clusterings. These common clusters are the result of the agreement between the evaluation measures we used and they define 6 stable Behavioural Groups of people. The 6 Behavioural Groups are indicated by the BG column in the tables and they identify the following behaviours:

- Group 1: People who have expensive houses, expensive assets and give a significant amount of their expenditure to pay back their debt and to other expenses.
- Group 2: People with low income and cheap cars but who spend a lot of money in travelling.
- Group 3: People with expensive houses who like to travel.
- Group 4: People with large income, not expensive houses who spend a lot in all categories, especially in the attributes priority, other and motoring and non expensive houses.
- Group 5: People who are characterized by large income and expensive assets.
- Group 6: People with large unsecured debts and lots of expenses in travelling.

Apart from the 6 stable groups of people we can identify through the clusters we can also see what characterizes the big distinct cluster that are depicted by turquoise color (cluster 1) in the biplots of Clara for stages A and B , by blue (cluster 3) in Kmeans for stage A and by red (cluster 2) in Kmeans for stage C. Inspecting the appropriate tables we can see that in 3 out of the 4 big clusters are characterized by the overexpression of travel attribute. Also 3 out of 4 clusters, but not the same 3, are characterized by low income. We can understand how these two expressions of these attributes can define a very large group of people and the fact that they don't agree in all 4 clusters may be a result of not perfect clustering. However, the big cluster appears to belong to the 2nd group we presented above in 3 of 4 cases.

Finally from the 6 different groups of people we identified from our clusterings and from the fact that a lot of the clusters in Kmeans, clusters 8-11, 12-13 in table 5, are characterized by the same attributes, validates our observation that the best number of clusters that contain the information we want and at the same time they are compact and well separated that is somewhere between 4 and 6 in the previous subsection.

### C. Selfish/ Non Selfish Characterization

In an attempt to associate the Behavioural Groups we discovered with aspects of Personality we processed the Selfishness ratings in order to produce a ranking of the expenditure attributes regarding how strongly are associated with Selfish Personality. In order to that, we calculated for each attribute the mean of the first score (Selfish) and the mean of the second score (Non Selfish) for all users. Then for each attribute we subtracted its Non Selfish mean from the Selfish mean and we got a weight of how strongly each attributes is defining Selfish personality. Finally the weights were normalized and ranked in descending order. The ranked attributes with their final weights can be seen in the table 6 and we can easily see that the attributes Leisure, Travel and Food are strongly associated with Selfish Personality while the Priority attribute is related with Non Selfish personality.

**Table 3 Expression of attributes in clusters produced in stage A**

| Kmeans | | | Clara | | |
|---|---|---|---|---|---|
| Cluster | Expression of Markers | BG | Cluster | Expression Of Markers | BG |
| 1 | hvalue+,mordebt+, carvalue+, income+ | 5 | 1 | income- | |
| 3 | travel+ | 2 | 2 | age+, carvalue+, hvalue+ | 5 |
| 4 | hvalue+,mordebt+ | | 3 | hvalue+, ndep+, travel+ | 3 |
| 5 | other+, motoring+, priority+,income+, udebt+, housing+, hvalue-, mordebt- | 4 | 4 | carvalue+, income+, motoring+ | 4 |
| 6 | hvalue+,mordebt+, income+, motoring+ | 1 | 5 | udebt+, travel+ | 6 |
| | | | 6 | motoring+, priority+, carvalue+, income+, hvalue+ | 1 |

**Table 4 Expression of attributes in clusters produced in stage B**

| Kmeans | | | Clara | | |
|---|---|---|---|---|---|
| Cluster | Expression of Markers | BG | Cluster | Expression Of Markers | BG |
| 2 | housing+, income+, other+, priority+ | 4 | 1 | age-, income-, carvalue-, travel+ | 2 |
| 3 | udebt+, other+, priority+, carvalue- | | 2 | hvalue+,travel+ | 3 |
| 5 | age+, carvalue+, income+ | 5 | 3 | age+, income+, carvalue+, other+, priority+ | 4 |
| 6 | age+, housing+, carvalue+, income+ | 5 | 4 | hvalue+, age+, income+, carvalue+ | 5 |
| 7 | hvalue+,travel+ | 3 | | | |
| 8 | hvalue+ | | | | |
| 9 | travel+, income-, carvalue- | 2 | | | |

**Table 5 Expression of attribuets in clusters produced in stage C**

| Kmeans | | | Clara | | |
|---|---|---|---|---|---|
| Cluster | Expression of Markers | BG | Cluster | Expression Of Markers | BG |
| 2 | income-, travel+, hvalue- | **2** | 1 | travel+, udebt+ | **6** |
| 3 | carvalue+, travel+, hvalue+ | **3** | 2 | income+,carvalue+, priority+,hvalue+ | 1 |
| 4 | travel+ | **2** | | | |
| 6 | carvalue+, travel+ | | | | |
| 7 | carvalue+, income+, priority+, hvalue+ | 1 | | | |
| 8-11 | carvalue+, income+, hvalue+ | 5 | | | |
| 12-13 | carvalue+, income+,other+, priority+, hvalue+ | 1 | | | |

Using this ranking we characterized the 5 of the 6 behavioural groups as Selfish and non − Selfish, based on the overexpression of expenditure types. With a closer look at the ranking of the expenditure attributes in table 2 we can see that travel is strongly associated with Selfish personality while priority is strongly associated with Non Selfish personality. Using that information we can conclude that all of the clusters that form the 2$^{nd}$, 3$^{rd}$ and 6$^{th}$ groups reveal a sort of Selfish behaviour while most of the clusters that form the 1$^{st}$ group (5/6) and 4$^{th}$ group (3/4) are characterized by Non Selfish behaviour. The Selfish Behavioural Groups are signified by the black background in BG column in tables 3, 4 and 5 while the Non Selfish by grey.

Our method of relating the cluster we extracted with Selfish and non − Selfish behaviour is based on the judgments of 52 students of Psychology that rated the 9 expenditure attributes of the dataset. This number is very small so as to indicate different spending behaviours and to reveal aspects of Personality. In addition to this 52 students is not a very big sample in order to guarantee the reliability of these ratings and our process of assessing and processing these ratins is not elaborate enough. For these reasons it is not safe to conclude that our findings reveal Selfish and non Selfish behaviour bat rather that they indicate signs of these behaviours. Nevertheless the significance of these findings lies in the fact that our method reveals a way to detect aspects of Personality in a socio economic dataset assuming that the assuming that the additional information someone can use is reliable and it is processed and applied in a careful and sophisticated manner.

**Table 6 Ranking of expenditure attributes towards selfishness**

| Expenditure Attribute | Selfish Weight | Expenditure Attribute | Selfish Weight |
|---|---|---|---|
| Leisure | 1.53 | Self Employed | -0.16 |
| Travel | 0.99 | Motoring | -0.20 |
| Food | 0.88 | Other | -0.42 |
| Clothing | 0.42 | Housing | -0.54 |
| | | Priority | -0.8 |

## VI. CONLCUSIONS AND FUTURE STEPS

In this paper we presented a way to extract behavioural groups from a socio economic dataset, such as CCCS, by using clustering methods. We linked these groups with Selfish and Non Selfish personalities by using the evaluation for the 10 expenditure types from 52 students of the Psychology department of the University of Nottingham.

Our findings demonstrate that it is possible to extract information regarding the Personality of individuals from similar datasets by using even simplistic data mining techniques. Despite its simplicity we hope that it will encourage the development of more sophisticated models to extract Personality Information. We believe that the information we extracted can be used to power up models that will lead us in interpreting complicated economic behaviours.

## ACKNOWLEDGMENT

We would like to thank John Gathergood, lecturer in school of economics for providing us the CCCS dataset and the 52 students of the Psychology Department for contributing voluntarly in scoring the expenditures.